# Separation of Water and Fat Magnetic Resonance Imaging Signals Using Deep Learning with Convolutional Neural Networks

James W. Goldfarb PhD [1]

[1] Department of Research and Education, Saint Francis Hospital, Roslyn, NY, USA

James W. Goldfarb Ph.D.

Department of Research and Education: DeMatteis MRI

St. Francis Hospital

100 Port Washington Boulevard

Roslyn, NY 11576

Email: James.Goldfarb.PhD@gmail.com

Phone: 516 622-4536

**Purpose:** A new method for magnetic resonance (MR) imaging water-fat separation using a convolutional neural network (ConvNet) and deep learning (DL) is presented. Feasibility of the method with complex and magnitude images is demonstrated with a series of patient studies and accuracy of predicted quantitative values is analyzed.

**Methods:** Water-fat separation of 1200 gradient-echo acquisitions from 90 imaging sessions (normal, acute and chronic myocardial infarction) was performed using a conventional model based method with modeling of $R_2^*$ and off-resonance and a multi-peak fat spectrum. A U-Net convolutional neural network for calculation of water-only, fat-only, $R_2^*$ and off-resonance images was trained with 900 gradient-echo bipolar gradient-echo acquisitions and the conventional method's results. 300 holdout acquisitions were used for visual and quantitative evaluation of the DL method. Multiple- and single-echo complex and magnitude input data algorithms were studied and compared to conventional extended echo modeling.

**Results:** The U-Net ConvNet was easily trained and provided water-fat separation results visually comparable to conventional methods. Myocardial fat deposition in chronic myocardial infarction and intramyocardial hemorrhage in acute myocardial infarction were well visualized in the DL results. Predicted values for $R_2^*$, off-resonance, water and fat signal intensities were well correlated with conventional model based water fat separation ($R^2>=0.97$, $p<0.001$). DL images had a 14% higher signal-to-noise ratio ($p<0.001$) when compared to the conventional method.

**Conclusion:** Deep learning utilizing ConvNets is a feasible method for MR water-fat separation imaging with complex, magnitude and single echo image data. A trained U-Net can be efficiently used for MR water-fat separation, providing results comparable to conventional model based methods.

# INTRODUCTION

Water-fat separation is a post-processing technique typically applied to multiple-echo magnetic resonance (MR) images to identify fat, provide images with fat suppression and quantify the amount and/or type of fat or lipids in specific tissues (1-4). This is most commonly accomplished with analytical multi-parameter modeling of the MR signal and individual pixel fitting. Multivariate modeling typically includes not only water and fat signals, but frequency offsets from main field inhomogeneities and $T_2^*$ relaxation. Additionally, multi-peak fat spectrum models (5) can be utilized for improved fitting (6). Complex or magnitude MR data are fit to these multi-parametric models with least squares or other advanced optimization techniques. These advanced models have led to a robust measurement of the proton fat density fat fraction (PDFF) for applications in the liver (7), bone marrow (8) and fat depots (9) as well as $R_2^*$ measurement without the effect of off-resonance (10). In contrast to early work (4,11), the process of water-fat separation has become somewhat complicated due to the specifics of the analytic model and optimization methods (12).

State-of-the-art chemical-shift signal models (12) include corrections for magnetic field inhomogeneities and $T_2^*$, but it is well known that there are other confounding factors (13-15). For example, $T_1$ recovery is rarely used in water-fat separation algorithms (16,17). Additionally, use of bipolar gradients when acquiring multiple echoes in a single repetition often complicates the reconstruction due to gradient delays and eddy currents (18). While analytical modeling improves the chemical-shift signal model representation, the additional variables can reduce the reconstruction speed and stability. Convolutional neural networks (ConvNets) (19,20) detect features or patterns in input data and combine these features for classification. Therefore, ConvNets offer a completely different post processing workflow via machine learning when compared to conventional water-fat separation model-based methods (4,21,22). Machine learning is a type of artificial intelligence that provides computers with the ability to learn a task without being explicitly programmed. It focuses on the development of computer programs that can change when exposed to new data. In this work, the task will be MR water-fat separation and the data will be a series of gradient-echo cardiovascular MR studies.

In this proof-of-concept study, we tested the feasibility of an end-to-end machine learning solution for water-fat separation in MR imaging. A U-Net ConvNet was trained using complex or magnitude-only gradient-echo cardiovascular images and conventional model-based water-fat reconstructions. Accuracy of quantitative values and image signal-to-noise were analyzed comparing input data (magnitude-only and complex) and the number input echoes. Additionally, the effects of a bipolar gradient acquisition and image fold-over artifacts on the machine learning approach and a conventional reconstruction were studied.

# METHODS

## Patient Population

This retrospective study used complex raw data from a database of prospective research cardiovascular studies. Some studies were part of prior publications (23-26). All subjects had signed an institutional review board (IRB) approved, Health Insurance Portability and Accountability Act compliant consent form prior to study initiation. Use for this study was approved by the local IRB by exemption. The database contained relevant studies from 66 subjects: 17 subjects without a cardiovascular history or complaints (normal controls), 13 acute myocardial infarction (MI) subjects at 3 days after infarction and 34 chronic MI subjects (>2 years after infarction). Additionally, there were 26 repeat sessions with the acute MI patients in the subacute phase of infarction. In total, 90 imaging session were included. The database consisted of 1204 acquisitions. On average, there were 13.5±3.0 acquisitions per subject. In this paper, the term acquisition is used to describe a single twelve gradient-echo acquisition which yields 12 images at increasing echo times.

## Imaging Protocol

All MR examinations were performed using one of two clinical 1.5 T imagers (Magnetom Avanto, Siemens Healthcare, Erlangen, Germany) with the subject in the supine position and standard ("tune-up") magnetic field shim. Images were acquired during suspended respiration at end-expiration with ECG gating and the standard 16 channel body phased array coil.

The pulse sequence was a dark blood double inversion recovery multiple spoiled gradient-echo sequence (1 slice per breathhold, repetition time = 20 ms; 12 echo times, 2.4 - 15.5 ms (1.2 ms spacing), flip angle = 20 degrees, bandwidth = 1860 Hz/pixel, in-plane spatial

resolution = 2.3 x 1.7 mm, slice thickness = 8 mm, flow compensation in read and slice). Every subject had acquisitions in 2-, 3- and 4- chamber long axis as well as short axis planes spanning the left ventricle. There were also several right ventricular outflow tract (RVOT) and axial planes in selected subjects. Multi-channel complex raw data was saved to the scanner's hard drive and archived.

## Image Reconstruction

Matlab R2016b (Mathworks, Waltham, MA) was used for inverse Fourier transformation, interpolation for proper rectangular field-of-view aspect ratios, coil combination and conventional (Conv.) water-fat separation. Zero-padding for two times interpolation and image cropping yielded an image matrix of 384x384. Coil combination was performed using locally relevant array correlation statistics (27). Coil combination yielded 12 complex images at consecutive echo times. The magnitude and real/imaginary images were saved to disk as and used as input for conventional and deep learning water-fat separation.

### Water-Fat Image Separation Conventional Method

Conventional water fat separation was performed via a multi-point fat-water separation with $R_2^*$ using a graph cut field map estimation algorithm (12,21) with the ISMRM water-fat Toolbox (28). Conventional processing of the bipolar acquisition was done separately using the even and odd echo images and results then averaged, producing water, fat, $R_2^*$ and off-resonance images for each acquisition.

The signal model of the gradient-echo acquisition at the nth echo time was:

$$I_n(\rho_W, \rho_F, R_2^*, \Delta f) = e^{-R_2^* TE_n} e^{j2\pi \Delta f\, TE_n} \left| \rho_W + \rho_F \sum_{p=1}^{P} \alpha_p\, e^{j2\pi f_{F,p}\, TE_n} \right|$$

where $\rho_W$ and $\rho_F$ are the water- and fat-only signals, $R_2^* = \frac{1}{T_2^*}$, $\Delta f$ is the off-resonance, $\alpha_p$ and $f_{F,p}$ are the amplitudes and frequencies for the multiple spectral peaks of fat relative to water (5) and $TE_n$ is the nth echo time.

All acquisitions were evaluated visually for artifacts and interesting cases for exclusion from the ConvNet training data. 120 images were identified with artifacts and excluded from training. Six imaging sessions, one normal, three chronic and two acute MI patients consisting of 80 acquisitions were also held out for method evaluation and not used in the training process. The images were used for ConvNet prediction and subsequent quantitative analysis of water- and fat-only images, $R_2$* and off-resonance. 1000 acquisitions were then available for training (900 fitting and 100 validation).

## Deep Learning

A U-Net (29) convolutional neural network was used for deep learning water-fat separation (Figure 1). Using all available data, the input to the training algorithm was 24 real and imaginary images from 12 echo times used as 24 channels of the ConvNet. Also investigated was the use of magnitude only inputs. Additionally, single echo acquisitions were simulated using only the first opposed-phase echo (TE=2.4 ms). The output of the ConvNet was four images (water only, fat only, $R_2$* and off-resonance) for complex input (Figure 1). For magnitude-only input, two ConvNets were used with the outputs of four images Similar to Figure 1) or two images (water only, fat only) (Supporting Figure S1). Individual input and output images were normalized by subtracting means and dividing by standard deviations estimated from the first 200 studies of the database. Output images were transformed back to units of Hz for $R_2$* and off-resonance and arbitrary units (a.u.) for water and fat images using the training means and standard deviations. A U-Net architecture was originally designed for fast and precise segmentation of images, but provides an architecture with multiresolution capability (29). The U-Net is defined by contraction (downsampling) followed by expansion (upsampling) as well as direct cross merging of features from the contraction path.   Light blue blocks in Figure 1 consist of convolutions, nonlinear rectified linear unit activation, dropout and max pooling. Green blocks consist of convolution, ReLU activation, dropout and upsampling. The final convolutional layer is performed with linear activation.

The implementation was realized using Keras 2.0 (30) and TensorFlow 1.2 (31) (both freely available software). Upon publication, the trained model and code generating the images in the Figures of this paper will be made publically available (github.com/jgoldfa/U-NetMriWaterFatSeparation). An EVGA GEFORCE GTX 1080 Ti graphics card was used for GPU accelerated calculations.  A mean-squared error metric was used. Training on n=900 [training set] (x12 echo-times) complex images (validation on n=100 images [validation set]) with 75 epochs was performed using the Adam optimizer with Nestrov momentum.  Data augmentation was performed by horizontal and vertical mirroring of the images.  Water and fat images from the data not used for training (n=300, [test set]) were "predicted" using the trained ConvNet.

Images were assessed for water-fat swap artifacts and visualization of ischemic cardiomyopathy (intramyocardial fat deposition and hemorrhage).  Region-of-interest (ROI) measurements of PDFF, $R_2^*$, off-resonance and signal-noise ratios (mean divided by the standard deviation of background noise) were performed in areas of myocardial infarction (fat deposition and intramyocardial hemorrhage) and normal septal myocardium in the hold out data.  Student's t-test and Pearson's correlation were performed to assess differences and the accuracy of the quantitative values in comparison to the conventional water-fat separation method.  A p-value < 0.05 was regarded as statistically significant.

# RESULTS

The U-Net ConvNet architecture learned the water-fat separation problem quickly (Supporting Figure S2) with a small amount of overfitting. Training past 20 epochs yielded only a modest reduction in mean-squared error. Training time per epoch was 400 seconds and total ConvNet training time was approximately 8 hours. Representative examples of DL water-fat prediction over a variety of anatomical slice planes in comparison to the conventional model based method are shown in Figures 2-3 and Supporting Figures S3 and S4. Water-fat separation was visually comparable between deep learning and the conventional model based method. Image resolution was equivalent between methods and well depicted intramyocardial fat deposition (Figure 2 and S3) in chronic myocardial infarction. Intramyocardial hemorrhage was well depicted in $R_2^*$ maps in areas of acute myocardial infarction, while water images showed homogeneous signal (Figure 3 and S4). Representative examples of magnitude-only and single-echo separation in the same individuals are shown in Figures 4 and 5. Magnitude only performed comparably well to complex separation. Single echo separation did not have the same high quality as 12 echo separation (Figure 5 and Supporting Figure S6), but performed well as a fat suppression method. Although not trained on images with fold-over artifacts, the deep learning method performed well in cases of severe fold-over Figure 6 . There were 26 acquisitions where the conventional model based method failed in inconsistencies of bipolar acquisition Supporting Figure S5. The DL method did not have this problem and performed well without bipolar acquisition artifacts.

There was an excellent correlation ($R^2>=0.97$, $p<0.001$) between the DL and conventional model based method (Figure 7) over a wide range of values for PDFF, $R_2^*$ and off-resonance with 12 complex echoes. Single-echo ConvNets provided good estimations of PDFF, and weak estimates of R2* and off-resonance (Supporting Material S6). The 12 echo magnitude ConvNet provided both excellent estimates of PDFF and R2*, but a weak estimate of off-resonance. Signal-to-noise ratio was consistently higher in DL (Figure 2 and Figure 3). In ROI analysis, it was 14% higher (SNR=8.4 vs 9.8), $p<0.001$.

## DISCUSSION

From seminal work to current reports, there have been numerous water-fat separation methods proposed. Most are based on chemical shift analytical modelling of the MR signal. Clear improvements in factors affecting accuracy have been made via inclusion of additional parameters and improved optimization allowing commercial implementations. Deep learning offers a paradigm shift for this classic problem in MR imaging. Neural networks are well known to be able to perform any computation (32), but ConvNets are a specialized type of neural networks and often used for visual interpretation of images, mimicking human visual processing.

In every day MR clinical practice, visual assessment of images is used to detect fat in MR images. Human readers are "trained" to detect fat based on shape, location and image intensity. The work here aims to replicate the human experience with artificial intelligence. The U-Net architecture is trained to detect features and combine these features for water-fat classification. Features can be detected in the image plane as well as in the echo time dimension. Other conventional water-fat separation (1) methods only use the echo time dimension via modelling of chemical shift.

In this proof-of-concept study, the quality of multiple echo DL water and fat only images was similar to the quality of a conventional method across several cardiovascular imaging planes. Additionally, correlation of quantitative predicted values was excellent; demonstrating that the deep learning approach is well suited to the water-fat separation problem. The U-Net architecture uses a multi resolution approach similar to other water fat separation approaches which use lower resolution phase and $T_2$* information for water-fat separation (33).  Additionally, the U-net architecture has found success in cardiac MR segmentation (34).  Also, ConvNets have been used for interpretation of pathological specimens (35) and medical images (36).  The DL method also provided good water-fat separation and estimates of R2* and off-resonance in a single-echo complex and magnitude-only scenarios.  The ConvNet in these scenarios is clearly learning features and predicting quantitative maps based on in plane information as multiple echoes are not available.  It is assumed that in the multiple echo situation, the ConvNet is using inplane features as well as chemical shift information.  The weak and moderate R2* and off-resonance correlations of are somewhat unexpected in the single-echo scenarios, but clearly demonstrate the power of a deep learning approach.  The image quality and quantitative accuracy of magnitude-only water-fat separation was not equivalent to a complex multi-echo separation, but could be used for fat suppression or identification of large fat.  It also could be retrospectively regularly utilized since only magnitude images are required.  Signal-to-noise measurements were performed to quantify a visual difference using a simplistic method.  An accurate measurement of signal-to-noise would include corrections for multiple receive coils and other factors.

The proposed solution is an end-to-end ConvNet solution without explicit model equations and markedly differs from another report (37) that utilizes ConvNets for water-fat separation.  Gong et al (37) first use a conventional model approach followed by a ConvNet to improve an initial model based  through improved fitting to the conventional model equations.  The method improves on Cui et al and reformulates surface fitting in residual space (38) as local convolutional steps (37).   Hence, the ConvNet in Gong et al, is not performing water-fat separation, rather improving a model based separation similar to a restorative ConvNet.

The results of this proof-of-concept study should be considered in light of some limitations. The study was limited to a single ConvNet architecture, cardiac MR images and a single MR protocol. There exist a number of ConvNet architectures such a RNN, VGG16, etc. that were not explored. Additionally, parameters of the ConvNet (number of features and layers) were not systematically tuned. A reduction in the number of features/model parameters was explored with a small increase in mean-squared error, which was not visually perceivable. Optimal tuning of ConvNets should be the subject of future work. The method was not tested on non-cardiac images such as the liver. Although not demonstrated, other non-cardiac structures were included in the cardiac images and it is expected that cardiac MR trained ConvNets would perform well in other anatomical areas. Adaption to flexible MR protocols (number and positions of echo times) was not explored, but it is expected that transfer learning (19,39) could be used to minimize future ConvNet training for these applications. Computation time was not compared between the methods, because TensorFlow utilizes parallel computations while the ISMRM Toolbox does not. Water-fat separation with DL was less than 250 ms per separation.

The feasibility of water-fat separation using an end-to-end ConvNet approach was demonstrated for complex, magnitude and single echo acquisitions. The ConvNet approach showed visually comparable images with slightly higher signal to noise in typical cardiac image planes. Quantitative image signal intensities, $R_2^*$ and off-resonance values showed excellent correlation with a conventional analytical model based method. The method worked well in areas with fold-over aliasing artifacts and a bipolar gradient acquisition. ConvNet based water-fat separation is a promising method capable of learning the water-fat separation problem with corrections for bipolar gradients, a multi-peak model, $R_2^*$ and off-resonance.

# Figure Captions

**Figure 1.** U-Net convolutional neural network used for water-fat separation (12 complex echoes). The input to the ConvNet is 24 images (12 echo times, real and imaginary components). The output is four images: water only, fat only, $R_2^*$ and off-resonance. Light blue blocks consist of 2D convolution, nonlinear activation, dropout and max pooling. Green blocks consist of 2D convolution, nonlinear activation, dropout and upsampling. Each step is labelled with the number of channels/features and image matrix size.

**Figure 2.** Water-fat separation performed comparatively well in multiple cardiac planes with deep learning (DL) when compared to the conventional (Conv.) method. A subject with and anterior wall chronic myocardial infarction shows fat deposition (red arrows) in water (hypo-intense) and fat only (hyper-intense) images. Although image resolution is equivalent between conventional and DL methods, the DL method consistently had higher signal-to-noise.

**Figure 3.** Comparison between deep learning (DL) and conventional (Conv.) water-fat separation in a subject with acute myocardial infarction. Water and fat images show excellent separation. $R_2^*$ images show an elevated $R_2^*$ (red arrows) due to intramyocardial hemorrhage.

**Figure 4.** Water-fat separation using a 12 echo magnitude-only input for the same patients from Figures 2 and 3. Fat deposition (red arrows, top row) and intramyocardial hemorrhage (red arrow, bottom row) are well depicted.

**Figure 5.** Water-fat separation using single-echo complex and magnitude-only inputs for the same patients in Figure 2 and 3. Image quality is not equivalent to 12 echo inputs, but overall separation is good and fat deposition (red arrows) is well depicted.

**Figure 6.** Deep learning (DL) performed well with image aliasing artifacts when compared to conventional (Conv.) separation, separating the aliased sections and underlying image sections.

**Figure 7.** Region-of-interest analysis showed an excellent correlation $R^2>=0.97$ between 12 echo complex deep learning and the conventional water-fat separation method for PDFF, $R_2^*$ and off-resonance quantitative values.

**Supporting Figure S1.** U-Net convolutional neural network used for single echo maggniture-only water-fat separation. The input to the ConvNet is a single magnitude image. The output is two images: water only, fat only. Light blue blocks consist of 2D convolution, nonlinear activation, dropout and max pooling. Green blocks consist of 2D convolution, nonlinear activation, dropout and upsampling. Each step is labelled with the number of channels/features and image matrix size.

**Supporting Figure S2.** Graph shows iterative training for 75 epochs of the U-Net convolutional neural network for water-fat separation. The mean-squared error decreases quickly for both training data (n=900, red) and validation data (n=100, black).

**Supporting Figure S3.** A subject with inferior wall chronic myocardial infarction shows fat deposition (red arrows) in water (hypo-intense) and fat only (hyper-intense) images. Although image resolution is equivalent between conventional and DL methods, the DL method consistently had higher signal-to-noise. An artifact from the CABG surgery sternal wires (white arrows) is visible.

**Supporting Figure S4.** Comparison between deep learning (DL) and conventional (Conv.) water-fat separation in a subject with an inferior wall acute myocardial infarction. Water and fat images in the two chamber view show excellent separation. $R_2^*$ images show an elevated $R_2^*$ (red arrows) due to intramyocardial hemorrhage.

**Supporting Figure S5.** DL water-fat separation worked well with the bipolar gradient echo acquisition. The conventional method sometimes confused water and fat and failed at water-fat separation and provided erroneous off-resonance maps (bottom row).

**Supporting Figure S6.** Region-of-interest analysis showed an excellent correlation $R^2 \geq 0.97$ using 12 echo complex deep learning. Single echo ConvNets provided good estimations of PDFF, and weak estimates of R2* and off-resonance. The 12 echo magnitude ConvNet provided both excellent estimates of PDFF and R2*, but a weak estimate of off-resonance.

# REFERENCES

1. Bley TA, Wieben O, Francois CJ, Brittain JH, Reeder SB. Fat and water magnetic resonance imaging. J Magn Reson Imaging 2010;31(1):4-18.
2. Ma J. Dixon techniques for water and fat imaging. J Magn Reson Imaging 2008;28(3):543-558.
3. Eggers H, Bornert P. Chemical shift encoding-based water-fat separation methods. J Magn Reson Imaging 2014;40(2):251-268.
4. Dixon WT. Simple proton spectroscopic imaging. Radiology 1984;153(1):189-194.
5. Hamilton G, Yokoo T, Bydder M, Cruite I, Schroeder ME, Sirlin CB, Middleton MS. In vivo characterization of the liver fat (1)H MR spectrum. NMR Biomed 2011;24(7):784-790.
6. Yu H, Shimakawa A, McKenzie CA, Brodsky E, Brittain JH, Reeder SB. Multiecho water-fat separation and simultaneous R2* estimation with multifrequency fat spectrum modeling. Magn Reson Med 2008;60(5):1122-1134.
7. Hines CD, Agni R, Roen C, Rowland I, Hernando D, Bultman E, Horng D, Yu H, Shimakawa A, Brittain JH, Reeder SB. Validation of MRI biomarkers of hepatic steatosis in the presence of iron overload in the ob/ob mouse. J Magn Reson Imaging 2012;35(4):844-851.
8. Gee CS, Nguyen JT, Marquez CJ, Heunis J, Lai A, Wyatt C, Han M, Kazakia G, Burghardt AJ, Karampinos DC, Carballido-Gamio J, Krug R. Validation of bone marrow fat quantification in the presence of trabecular bone using MRI. J Magn Reson Imaging 2015;42(2):539-544.
9. Hu HH, Hines CD, Smith DL, Jr., Reeder SB. Variations in T(2)* and fat content of murine brown and white adipose tissues by chemical-shift MRI. Magn Reson Imaging 2012;30(3):323-329.
10. Hernando D, Vigen KK, Shimakawa A, Reeder SB. R*(2) mapping in the presence of macroscopic B(0) field variations. Magn Reson Med 2012;68(3):830-840.
11. Glover GH, Schneider E. Three-point Dixon technique for true water/fat decomposition with B0 inhomogeneity correction. Magn Reson Med 1991;18(2):371-383.
12. Hernando D, Liang ZP, Kellman P. Chemical shift-based water/fat separation: a comparison of signal models. Magn Reson Med 2010;64(3):811-822.
13. Hernando D, Sharma SD, Kramer H, Reeder SB. On the confounding effect of temperature on chemical shift-encoded fat quantification. Magn Reson Med 2014;72(2):464-470.
14. Bydder M, Yokoo T, Hamilton G, Middleton MS, Chavez AD, Schwimmer JB, Lavine JE, Sirlin CB. Relaxation effects in the quantification of fat using gradient echo imaging. Magn Reson Imaging 2008;26(3):347-359.
15. Peterson P, Svensson J, Mansson S. Relaxation effects in MRI-based quantification of fat content and fatty acid composition. Magn Reson Med 2014;72(5):1320-1329.
16. Hines CD, Yu H, Shimakawa A, McKenzie CA, Brittain JH, Reeder SB. T1 independent, T2* corrected MRI with accurate spectral modeling for quantification of fat: validation in a fat-water-SPIO phantom. J Magn Reson Imaging 2009;30(5):1215-1222.
17. Meisamy S, Hines CD, Hamilton G, Sirlin CB, McKenzie CA, Yu H, Brittain JH, Reeder SB. Quantification of hepatic steatosis with T1-independent, T2-corrected MR imaging with spectral modeling of fat: blinded comparison with MR spectroscopy. Radiology 2011;258(3):767-775.
18. Lu W, Yu H, Shimakawa A, Alley M, Reeder SB, Hargreaves BA. Water-fat separation with bipolar multiecho sequences. Magn Reson Med 2008;60(1):198-209.
19. Shin H-C, Roth HR, Gao M, Lu L, Xu Z, Nogues I, Yao J, Mollura D, Summers RM.

Deep convolutional neural networks for computer-aided detection: CNN architectures, dataset characteristics and transfer learning. IEEE transactions on medical imaging 2016;35(5):1285-1298.
20. Schmidhuber J. Deep learning in neural networks: An overview. Neural networks 2015;61:85-117 %@ 0893-6080.
21. Hernando D, Kellman P, Haldar JP, Liang ZP. Robust water/fat separation in the presence of large field inhomogeneities using a graph cut algorithm. Magn Reson Med 2010;63(1):79-90.
22. Ma J. Breath-hold water and fat imaging using a dual-echo two-point Dixon technique with an efficient and robust phase-correction algorithm. Magn Reson Med 2004;52(2):415-419.
23. Goldfarb JW, Hasan U. Imaging of reperfused intramyocardial hemorrhage with cardiovascular magnetic resonance susceptibility weighted imaging (SWI). PLoS One 2015;10(4):e0123560.
24. Goldfarb JW, Hasan U, Zhao W, Han J. Magnetic resonance susceptibility weighted phase imaging for the assessment of reperfusion intramyocardial hemorrhage. Magn Reson Med 2014;71(3):1210-1220.
25. Goldfarb JW, Roth M, Han J. Myocardial fat deposition after left ventricular myocardial infarction: assessment by using MR water-fat separation imaging. Radiology 2009;253(1):65-73.
26. Goldfarb JW. Fat-water separated delayed hyperenhanced myocardial infarct imaging. Magn Reson Med 2008;60(3):503-509.
27. Walsh DO, Gmitro AF, Marcellin MW. Adaptive reconstruction of phased array MR imagery. Magn Reson Med 2000;43(5):682-690.
28. Hu HH, Bornert P, Hernando D, Kellman P, Ma J, Reeder S, Sirlin C. ISMRM workshop on fat-water separation: insights, applications and progress in MRI. Magn Reson Med 2012;68(2):378-388.
29. Ronneberger O, Fischer P, Brox T. U-net: Convolutional networks for biomedical image segmentation. In 2015. Springer. p 234-241.
30. Chollet F, others a. Keras. GitHub; 2015.
31. Abadi M, Barham P, Chen J, Chen Z, Davis A, Dean J, Devin M, Ghemawat S, Irving G, Isard M. TensorFlow: A system for large-scale machine learning. In 2016.
32. Csáji BC. Approximation with artificial neural networks. Faculty of Sciences, Etvs Lornd University, Hungary 2001;24:48.
33. Lu W, Hargreaves BA. Multiresolution field map estimation using golden section search for water-fat separation. Magn Reson Med 2008;60(1):236-244.
34. Goldfarb JW, Cao JJ, DeWit J. Automated Left Ventricular Volumetric Quantitation from Short-axis CMR Images with Machine Learning using a Deep Convolutional Neural Network. In Proceedings of the 25th Annual Meeting of ISMRM; 2017; Honolulu, HI. p 634.
35. Litjens G, Sanchez CI, Timofeeva N, Hermsen M, Nagtegaal I, Kovacs I, Hulsbergen-van de Kaa C, Bult P, van Ginneken B, van der Laak J. Deep learning as a tool for increased accuracy and efficiency of histopathological diagnosis. Sci Rep 2016;6:26286.
36. Litjens GJ, Elliott R, Shih NN, Feldman MD, Kobus T, Hulsbergen-van de Kaa C, Barentsz JO, Huisman HJ, Madabhushi A. Computer-extracted Features Can Distinguish Noncancerous Confounding Disease from Prostatic Adenocarcinoma at Multiparametric MR Imaging. Radiology 2016;278(1):135-145.
37. Gong E, Zaharchuk G, Pauly J. Improved Multi-echo Water-fat Separation Using Deep Learning. In Proceedings of the 25th Annual Meeting of ISMRM; 2017; Honolulu, HI. p 5657.
38. Cui C, Wu X, Newell JD, Jacob M. Fat water decomposition using globally optimal

        surface estimation (GOOSE) algorithm. Magn Reson Med 2015;73(3):1289-1299.
39. Ng H-W, Nguyen VD, Vonikakis V, Winkler S. Deep learning for emotion recognition on small datasets using transfer learning. In 2015. ACM. p 443-449.

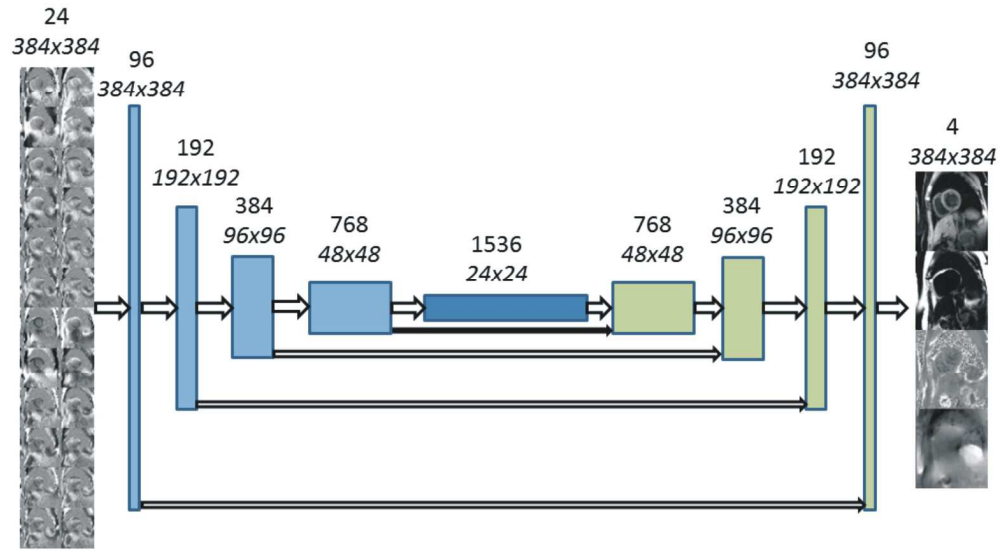

Figure 1. U-Net convolutional neural network used for water-fat separation (12 complex echoes).  The input to the ConvNet is 24 images (12 echo times, real and imaginary components). The output is four images: water only, fat only, R2* and off-resonance.  Light blue blocks consist of 2D convolution, nonlinear activation, dropout and max pooling.  Green blocks consist of 2D convolution, nonlinear activation, dropout and upsampling. Each step is labelled with the number of channels/features and image matrix size.

1269x717mm (72 x 72 DPI)

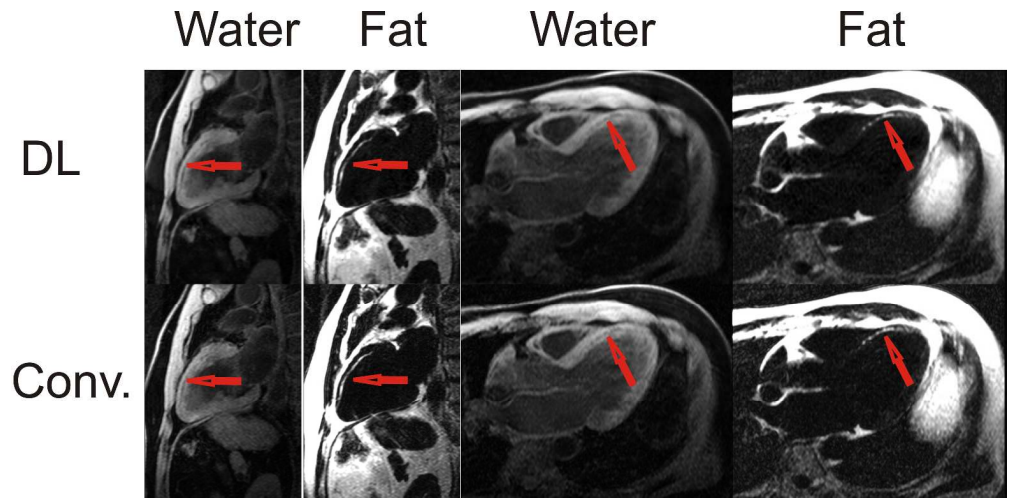

Figure 2. Water-fat separation performed comparatively well in multiple cardiac planes with deep learning (DL) when compared to the conventional (Conv.) method. A subject with and anterior wall chronic myocardial infarction shows fat deposition (red arrows) in water (hypo-intense) and fat only (hyper-intense) images. Although image resolution is equivalent between conventional and DL methods, the DL method consistently had higher signal-to-noise.

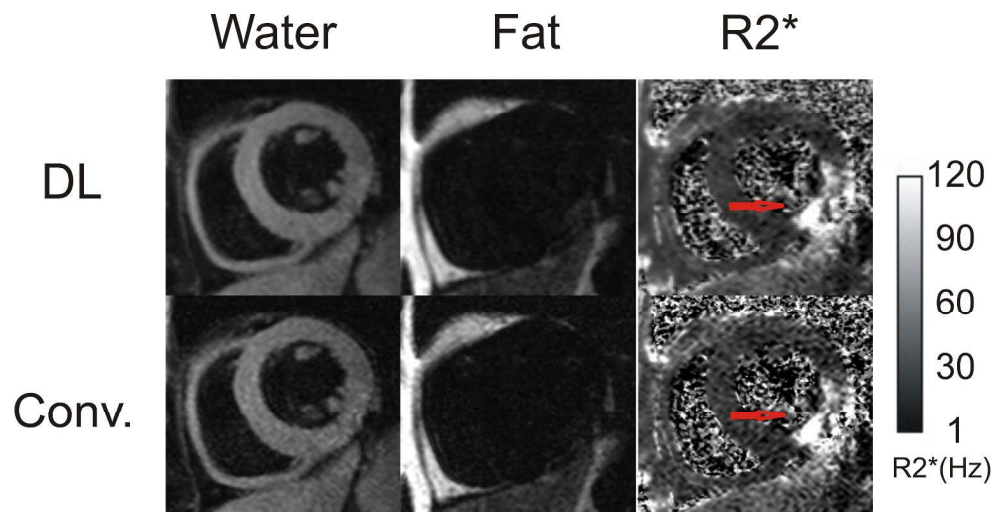

Figure 3. Comparison between deep learning (DL) and conventional (Conv.) water-fat separation in a subject with acute myocardial infarction. Water and fat images show excellent separation. R2* images show an elevated R2* (red arrows) due to intramyocardial hemorrhage.

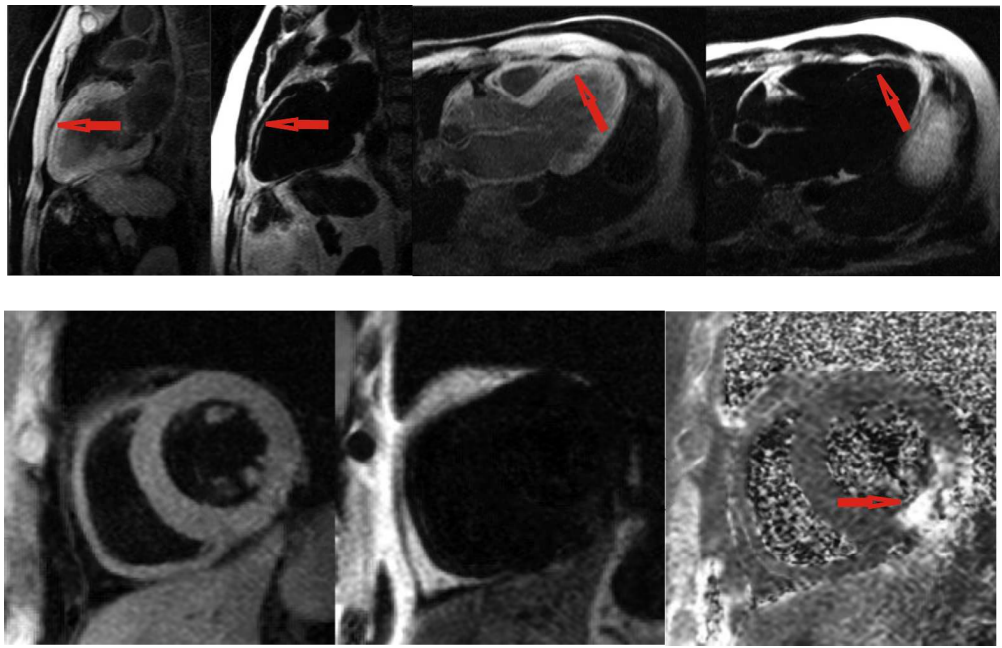

Figure 4.  Water-fat separation using a 12 echo magnitude-only input for the same patients from Figures 2 and 3.  Fat deposition (red arrows, top row) and intramyocardial hemorrhage (red arrow, bottom row) are well depicted.

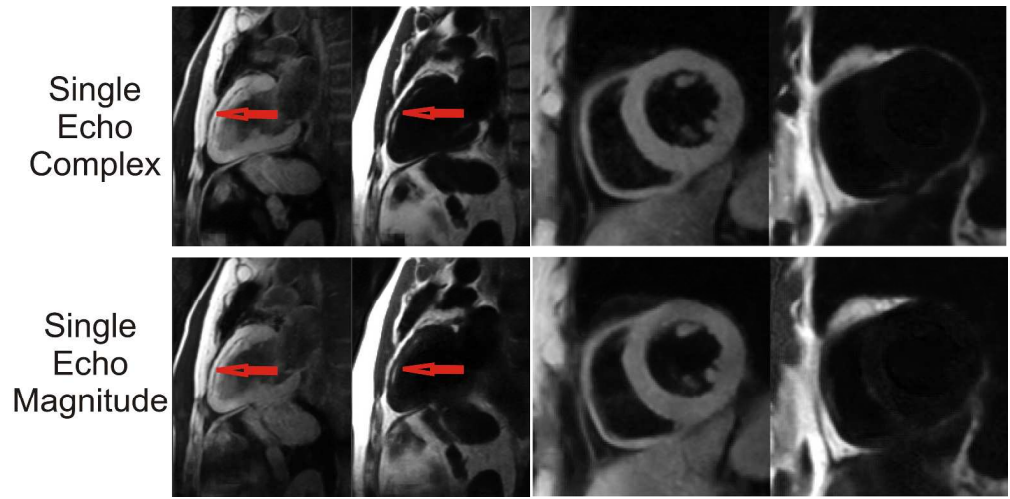

Figure 5. Water-fat separation using single-echo complex and magnitude-only inputs for the same patients in Figure 2 and 3. Image quality is not equivalent to 12 echo inputs, but overall separation is good and fat deposition (red arrows) is well depicted.

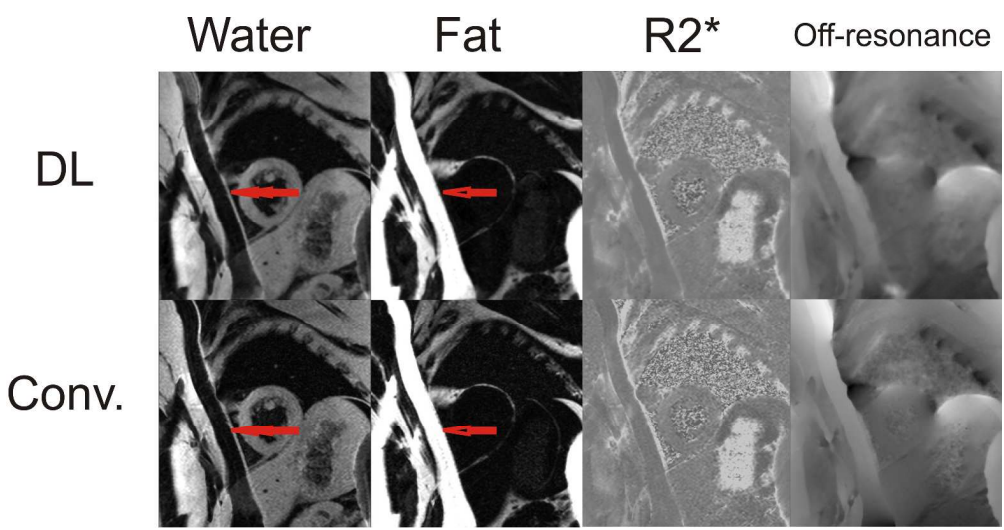

Figure 6. Deep learning (DL) performed well with image aliasing artifacts when compared to conventional (Conv.) separation, separating the aliased sections and underlying image sections.

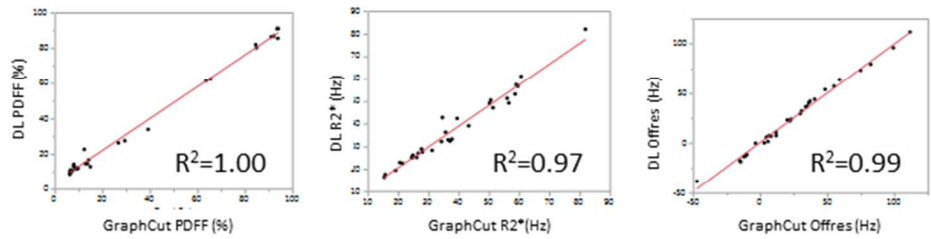

Figure 7.  Region-of-interest analysis showed an excellent correlation R2>=0.97 between 12 echo complex deep learning and the conventional water-fat separation method for PDFF, R2* and off-resonance quantitative values.

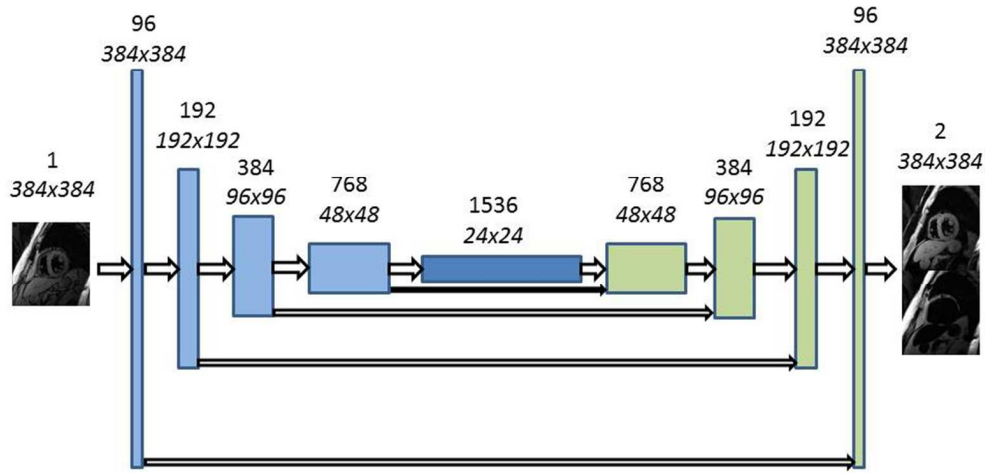

Supporting Figure S1. U-Net convolutional neural network used for single echo maggniture-only water-fat separation. The input to the ConvNet is a single magnitude image. The output is two images: water only, fat only. Light blue blocks consist of 2D convolution, nonlinear activation, dropout and max pooling. Green blocks consist of 2D convolution, nonlinear activation, dropout and upsampling. Each step is labelled with the number of channels/features and image matrix size.

254x190mm (96 x 96 DPI)

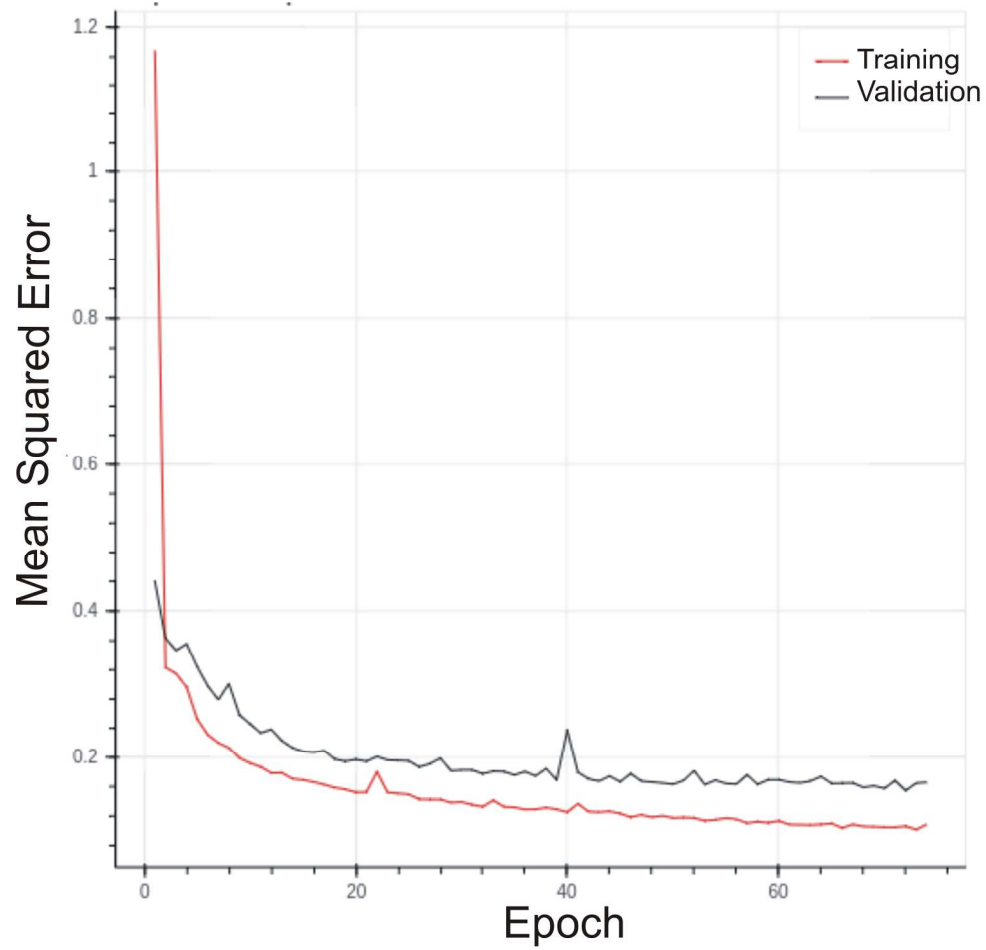

Supporting Figure S2.  Graph shows iterative training for 75 epochs of the U-Net convolutional neural network for water-fat separation.  The mean-squared error decreases quickly for both training data (n=900, red) and validation data (n=100, black).

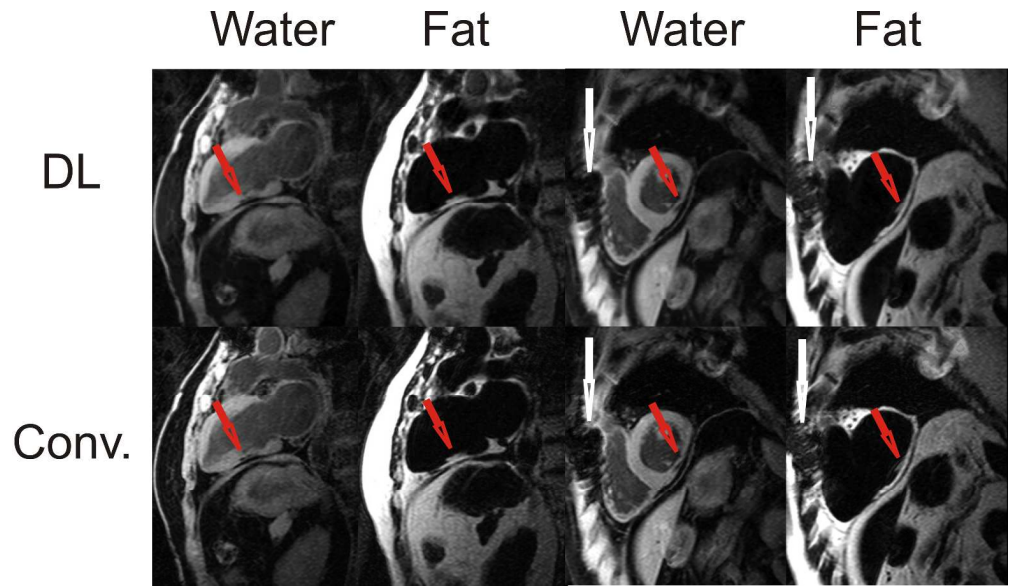

Supporting Figure S3. A subject with inferior wall chronic myocardial infarction shows fat deposition (red arrows) in water (hypo-intense) and fat only (hyper-intense) images. Although image resolution is equivalent between conventional and DL methods, the DL method consistently had higher signal-to-noise. An artifact from the CABG surgery sternal wires (white arrows) is visible.

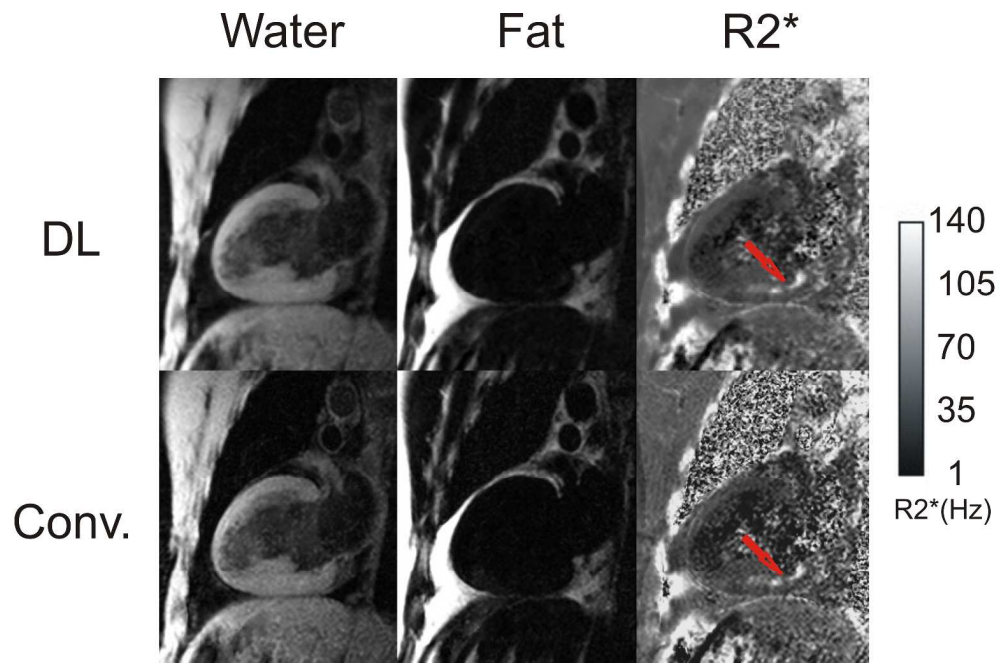

Supporting Figure S4. Comparison between deep learning (DL) and conventional (Conv.) water-fat separation in a subject with an inferior wall acute myocardial infarction. Water and fat images in the two chamber view show excellent separation. R2* images show an elevated R2* (red arrows) due to intramyocardial hemorrhage.

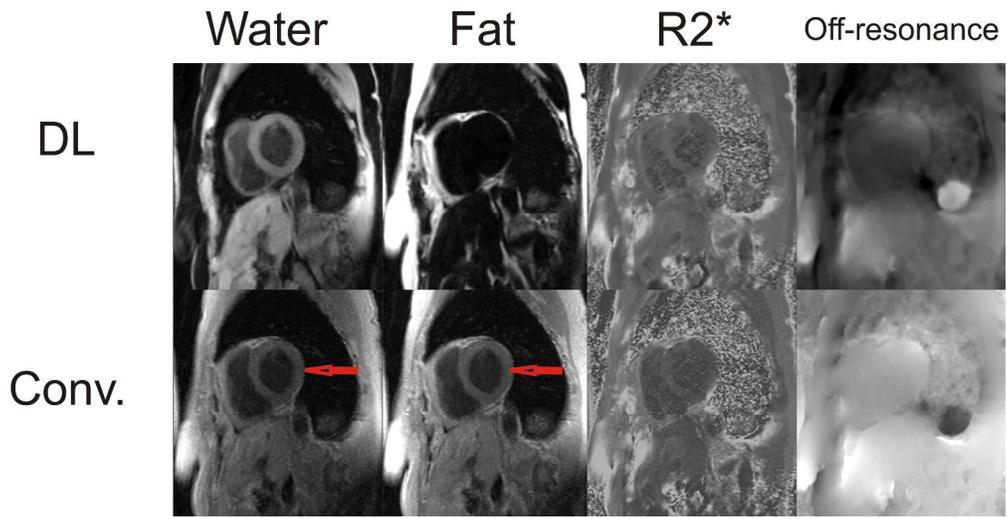

Supporting Figure S5.  DL water-fat separation worked well with the bipolar gradient echo acquisition.  The conventional method sometimes confused water and fat and failed at water-fat separation and provided erroneous off-resonance maps (bottom row).

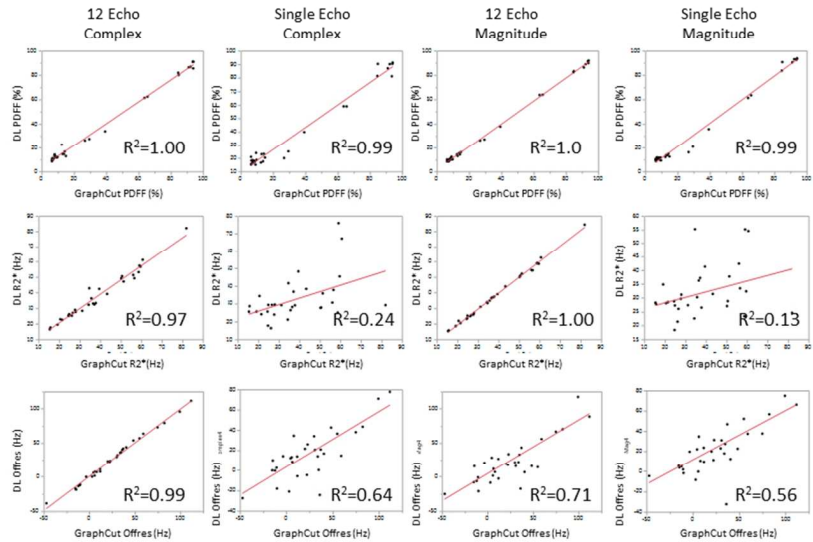

Supporting Figure S6. Region-of-interest analysis showed an excellent correlation R2>=0.97 using 12 echo complex deep learning. Single echo ConvNets provided good estimations of PDFF, and weak estimates of R2* and off-resonance. The 12 echo magnitude ConvNet provided both excellent estimates of PDFF and R2*, but a weak estimate of off-resonance.

254x190mm (96 x 96 DPI)